\documentclass[11pt,a4paper]{article}
\usepackage[T1]{fontenc}
\usepackage{authblk}
\usepackage{graphicx}
\usepackage{xcolor}
\usepackage{microtype}
\usepackage[
    colorlinks=true,
    linkcolor=blue,
    anchorcolor=blue,
    citecolor=blue,
    filecolor=blue,
    urlcolor=blue
]{hyperref}
\usepackage{booktabs}
\usepackage{multirow}
\usepackage{amsmath}
\usepackage{bm}
\usepackage[capitalize]{cleveref}
\usepackage{xurl}

\usepackage{acronym}
\newacro{cnt}[\textsc{cnt}]{Commentarii Notarum Tironianarum}
\newacro{hd-cnn}[\textsc{hd-cnn}]{Hierarchical Deep Convolutional Neural Network}
\newacro{snt}[\textsc{snt}]{Supertextus Notarum Tironianarum} %
\newacro{htr}[\textsc{htr}]{Handwritten Text Recognition}
\newacro{jem}[\textsc{jem}]{joint energy-based model}
\newacro{ocr}[\textsc{ocr}]{Optical Character Recognition}
\newacro{ssl}[\textsc{ssl}]{Self-Supervised Learning}
\newacro{ebm}[\textsc{ebm}]{energy-based model}
\newacro{ai}[\textsc{ai}]{artificial intelligence}
\newacro{msi}[\textsc{msi}]{Multispectral Imaging}
\newacro{gan}[\textsc{gan}]{Generative Adversarial Network}
\newacro{dh}[\textsc{dh}]{Digital Humanities}
\newacro{icdar}[\textsc{icdar}]{International Conference on Document Analysis and Recognition}
\newacro{cnn}[\textsc{cnn}]{Convolutional Neural Network}
\newacro{vit}[\textsc{vit}]{Vision Transformer}
\newacro{swin}[\textsc{swin}]{Shifted Window Transformer}
\newacro{resnet}[\textsc{resnet}]{Residual Network}
\newacro{cs}[\textsc{cs}]{Computer Science}
\newacro{b-cnn}[\textsc{b-cnn}]{Branch Convolutional Neural Network}
\newacro{tree-cnn}[\textsc{tree-cnn}]{Tree-structured Convolutional Neural Network}
\newacro{knn}[\textsc{knn}]{K-nearest-neighbor}

\begin{document}

\title{Evaluating Hierarchy-Aware Deep Learning for the Recognition of Tironian Notes}
\author[1]{Yule Kang}
\author[1]{Thomas Gorges\,\href{https://orcid.org/0009-0007-0573-0992}{(ORCID)}}
\author[2]{Janne van der Loop\,\href{https://orcid.org/0000-0001-7486-669X}{(ORCID)}}
\author[2]{Franziska Marske\,\href{https://orcid.org/0009-0006-2649-9948}{(ORCID)}}
\author[3]{Nikolaus Weichselbaumer\,\href{https://orcid.org/0000-0002-8065-0390}{(ORCID)}}
\author[4]{Tino Licht}
\author[1]{Vincent Christlein\,\href{https://orcid.org/0000-0003-0455-3799}{(ORCID)}}
\date{}
\affil[1]{Pattern Recognition Lab, Friedrich-Alexander-Universität Erlangen-Nürnberg, Germany\\
\texttt{yule.kang@fau.de, thomas.gorges@fau.de, vincent.christlein@fau.de}}

\affil[2]{Buchwissenschaft, Johannes Gutenberg-Universität Mainz, Germany\\
\texttt{jannevanderloop@uni-mainz.de, fmarske@students.uni-mainz.de}}

\affil[3]{Abteilung Handschriften und Historische Drucke, Staatsbibliothek zu Berlin, Germany\\
\texttt{Nikolaus.Weichselbaumer@sbb.spk-berlin.de}}

\affil[4]{Lateinische Philologie des Mittelalters und der Neuzeit am Historischen Seminar, Universität Heidelberg, Germany\\
\texttt{Tino.Licht@urz.uni-heidelberg.de}}

\maketitle              %
\begin{abstract}
Tironian notes are generally regarded as the first Latin shorthand system and are notable for their large, fine-grained symbol inventory.
Their high visual similarity and large class set make manual reading time-consuming, leaving manuscripts that contain Tironian notes inaccessible to many researchers.
Automatic recognition is also challenging because models must distinguish subtle differences in stroke shape and sign structure while realistic training data remain scarce.
However, standard flat classifiers do not explicitly use visual or structural relations between related signs.
This paper investigates whether structural relationships between Tironian notes can support automatic recognition.
We use the \ac{snt} by Martin Hellmann, which provides idealized sign forms and a hierarchical organization of Tironian notes.
We compare flat ResNet18, ConvNeXt, \ac{swin}, and \ac{vit} classifiers with \ac{hd-cnn}-style coarse-to-fine models and hierarchy-aware routing models based on visual class cleaning and similarity-based re-clustering.
The models are evaluated on handwritten samples and manuscript-domain samples from Vergilius Turonensis, both with and without limited few-shot adaptation to the manuscript domain.
The results show that the relative performance of flat and hierarchical models depends on adaptation.
On Vergilius Turonensis, \ac{hd-cnn} achieves the best non-adapted Top-1 result with 45.43\,\%, while flat classification reaches the best Top-1 result after few-shot adaptation with 82.09\,\%.
Overall, the results indicate that hierarchical structure can support Tironian note recognition, especially under non-adapted conditions. %


\end{abstract}

\section{Introduction}
Tironian notes are an ancient Latin shorthand system that was widely used in medieval manuscripts in the early Middle Ages.
Unlike alphabetic writing systems, individual Tironian symbols typically represent complete words, word parts, or phrases rather than single letters~\cite{schmitz1893commentarii}.
Because of this compact and specialized notation, manuscripts containing Tironian notes remain inaccessible to many researchers, as their interpretation typically requires highly specialized expertise~\cite{pearse_cnt}.

Historical reference materials exist, including the \ac{cnt} as an index of Tironian notes~\cite{schmitz1893commentarii} and the more recent \acf{snt} by Martin Hellmann, a digital hypertext lexicon in which Tironian symbols are organized hierarchically~\cite{hellmann_martinellus}.
This structure provides a basis for moving beyond flat classification and investigating whether visual and structural relationships between symbols can support automatic recognition.

Automatic recognition remains challenging under realistic manuscript conditions.
The symbol inventory is large and fine-grained; many symbols are visually similar and differ only in subtle local stroke patterns or small structural variations.
Furthermore, there is a strong visual gap between clean standard symbols, modern handwritten samples, and historical manuscript crops.
Annotated manuscript-domain data are scarce, and historical manuscript images may exhibit background texture, degraded strokes, low contrast, and incomplete local details.
As a result, Tironian note recognition is not only a fine-grained image classification problem, but also a low-resource and domain-adaptation problem~\cite{lombardi2020deep}.

Standard deep learning classifiers typically treat this task as flat multi-class classification, where each input image is mapped directly to one label from the full symbol inventory~\cite{he2016deep,liu2022convnet,liu2021swin,dosovitskiy2020image}.
This provides a strong baseline, but relations between visually or structurally related symbols are not explicitly encoded in the flat classification objective.
Coarse-to-fine and hierarchy-aware routing strategies offer an alternative by organizing the label space into broader groups before predicting fine-grained classes~\cite{silla2011survey,yan2015hd}.
In this study, both hierarchical strategies, Coarse-to-fine and hierarchy-aware routing, are built from visually generated clustering trees; each of them uses the tree structure differently. %
The \acf{hd-cnn} model~\cite{yan2015hd} uses only the lowest two levels of the generated tree: the leaf labels define the fine classes, while their immediate parent nodes define the coarse groups.
The hierarchy-aware routing model uses a deeper part of the generated tree and performs prediction as a sequence of local routing decisions through multiple hierarchy levels.

This paper evaluates three recognition strategies for Tironian notes: flat classification, \ac{hd-cnn} coarse-to-fine classification, and hierarchy-aware routing classification.
The experiments investigate whether structured approaches can support recognition under low-resource, manuscript-domain conditions, while also accounting for the potential instability introduced by additional routing decisions.
For the evaluation, clean standard symbols are used, along with augmented training data, modern handwritten notes, and expert-cropped manuscript symbols from Vergilius Turonensis\cite{ecodices_vergilius}.
In addition, we study the effect of a limited few-shot adaptation to the manuscript domain.
The comparison focuses on recognition and Top-k performance.

\section{Related Work}
Deep learning has been widely used in visual recognition and historical document analysis, where \ac{cnn}s learn local visual patterns such as edges, strokes, curves, and shape fragments~\cite{lecun2002gradient,lecun2015deep}.
For symbol recognition, these local patterns often determine the differences between visually similar classes.
Modern image classification architectures extend this principle in different ways.
\ac{resnet} uses residual connections to enable the training of deeper convolutional networks~\cite{he2016deep}, while ConvNeXt updates convolutional architectures with design choices inspired by recent vision models~\cite{liu2022convnet}.
Transformer-based models such as \ac{vit} and \ac{swin} use self-attention to model relationships between image regions; \ac{swin} further combines local shifted windows with a hierarchical feature design~\cite{dosovitskiy2020image,liu2021swin}.

Tironian note recognition follows the pattern of historical document recognition, which often involves limited annotated data, domain shifts, and degraded image quality~\cite{lombardi2020deep}, and additionally works from clean standard forms, modern handwritten samples, as well as from historical manuscript crops that differ in background, contrast, stroke quality, and writing style.
Data augmentation is commonly used under such low-resource conditions to increase training-data variability~\cite{shorten2019survey}.
For handwritten symbol recognition, augmentation must preserve stroke structure because excessive geometric transformations can alter a symbol's visual identity.
Chen~\textit{et al.}~\cite{chen2022script} propose a script-level augmentation method for few-shot handwritten text recognition, where generated variants preserve writing structure, which is relevant for Tironian notes as well, where small stroke-level modifications can heavily change the symbol's meaning.

When labels can be organized into groups or multi-level structures, hierarchical classification provides an alternative to flat prediction~\cite{silla2011survey}.
Instead of predicting directly over all labels, hierarchical methods divide recognition into broader and finer decisions.
Combining a shared feature extractor, a coarse classifier, and several fine-level specialist classifiers, \ac{hd-cnn} serves as a representative coarse-to-fine visual recognition model where fine predictions are made within coarse category groups~\cite{yan2015hd}.
Other neural hierarchical models use label structure differently.
Organizing classification through a tree-structured model, \ac{tree-cnn} was introduced in the context of incremental learning~\cite{roy2020tree}, whereas \ac{b-cnn} utilizes multiple classification branches corresponding to different hierarchy levels within a shared convolutional network~\cite{zhu2017b}.
These approaches represent different ways of integrating label structure into neural recognition.

Not only does the performance of hierarchical classification depend on the model architecture, but it also depends on the label hierarchy.
Label tree learning studies how to arrange class labels into tree structures for efficient or structured prediction~\cite{deng2011fast}.
Extreme classification methods such as Parabel and Bonsai also use label trees, with different choices of depth, balance, and tree diversity~\cite{prabhu2018parabel,khandagale2020bonsai}.
For visually grounded hierarchy construction, agglomerative clustering can build nested groups from feature distances~\cite{mullner2011modern,rokach2005clustering}.
Hierarchical inference methods further address how local tree decisions are combined into final predictions~\cite{liu2013probabilistic,sun2013find}.
Building on this work, this paper compares flat classifiers, \ac{hd-cnn}-style two-level coarse-to-fine models, and hierarchy-aware routing models based on visually generated clustering trees.

\section{Data}
\begin{figure}[t]
\centering
\includegraphics[width=0.8\textwidth]{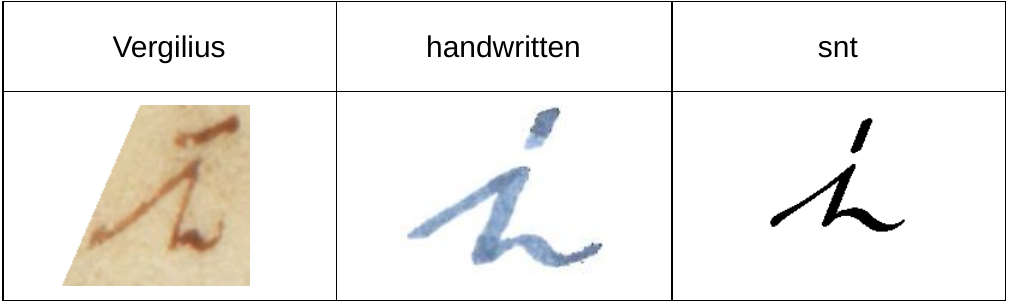}
\caption{Examples of visually corresponding Tironian symbols from the three main visual domains used in this paper: historical manuscript crops from Vergilius Turonensis~\cite{ecodices_vergilius}, modern handwritten samples from the internal handwritten notes dataset created by \textsc{jgu} Mainz and Heidelberg University, and clean standard forms from the \ac{snt}~\cite{hellmann_martinellus}.}
\label{fig:dataset-examples}
\end{figure}

The evaluation covers three main visual domains for Tironian note recognition: clean standard symbols from \ac{snt}~\cite{hellmann_martinellus}, modern handwritten samples, and cropped historical Tironian notes from the Vergilius Turonensis manuscript~\cite{ecodices_vergilius}.
These sources differ in visual appearance, class coverage, origin, and role in the evaluation pipeline.
As additional training data, augmented variants of the \ac{snt} standard symbols are used.
\Cref{fig:dataset-examples} shows representative examples from the three visual domains.

\Cref{tab:dataset-overview} summarizes the image data sources used in the experiments.
\ac{snt} provides clean standard symbol images and metadata~\cite{hellmann_martinellus}. These standard images provide broad class coverage and are used for training and for constructing visual hierarchies.
Since many classes contain a small number of standard samples or even only one, augmented \ac{snt} variants are generated to increase the amount of training data.
These generated samples are used for training and validation only and are not treated as an independent evaluation domain.

\begin{table}[tb]
\setlength{\tabcolsep}{6pt}
\caption{Overview of the data sources used in this paper.}
\label{tab:dataset-overview}
\centering
\begin{tabular}{p{2.8cm}p{5.2cm}rr}
\toprule
\textbf{Source} & \textbf{Description} & \textbf{Classes} & \textbf{Samples} \\
\midrule
\ac{snt} & Standard symbol images, metadata, and reference hierarchy & 14,445 & 15,898 \\
Augmented \ac{snt} & Synthetic variants of \ac{snt} symbols & 14,445& 254,368 \\
Handwritten notes & Modern handwritten samples from \textsc{jgu} Mainz and Heidelberg University & 310 & 57,179 \\
Vergilius Turonensis & Cropped manuscript symbols from the \mbox{e-codices} facsimile & 142 & 691\\
\bottomrule
\end{tabular}
\end{table}

In this paper, the hierarchical organization of Tironian symbols of the \ac{snt}~\cite{hellmann_martinellus} is represented in a structured form and retained as a reference hierarchy.
It is, however, not used directly as the training hierarchy, since recognition strategies that require hierarchical grouping use visually generated clustering trees.
This separates the lexicon-based organization of the \ac{snt} from the visual grouping for model evaluation.

The handwritten notes dataset is an internal dataset and was created by \textsc{jgu} Mainz and Heidelberg University.
The modern handwritten reproductions of Tironian symbols on a clean background introduce variation in stroke shape, stroke thickness, and spatial arrangement, while avoiding the more pronounced degradation found in historical manuscript crops.
The modern dataset covers only a subset of the full \ac{snt} label inventory and mainly contains the more frequent symbols.
It is used for training and for controlled-domain evaluation.

\begin{figure}[t]
\centering
\includegraphics[width=0.8\textwidth]{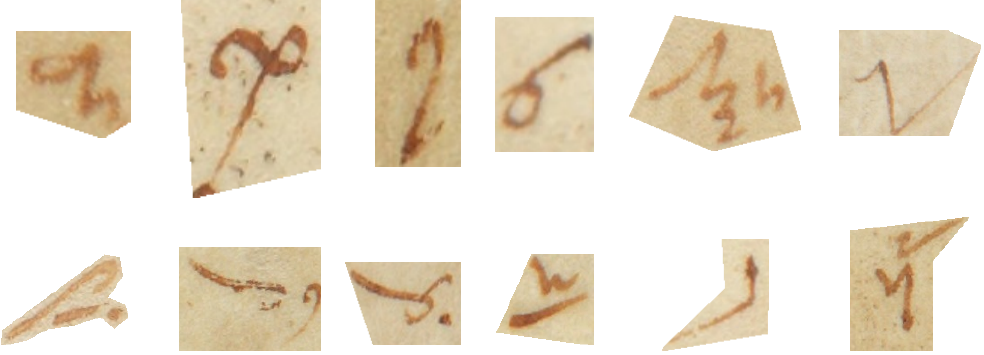}
\caption{Representative cropped Tironian note samples from Vergilius Turonensis~\cite{ecodices_vergilius}.}
\label{fig:vergilius-example}
\end{figure}

The Vergilius Turonensis dataset contains cropped Tironian symbols from historical manuscript pages.
The original digitized manuscript is available through e-codices~\cite{ecodices_vergilius}, while the cropped dataset used in this paper was created before this study by \textsc{jgu} Mainz and Heidelberg University.
Compared with the standard and handwritten data, these crops contain background texture, uneven contrast, ink variation, degraded strokes, and incomplete local details.
Representative samples are shown in \cref{fig:vergilius-example}.
This dataset was used for manuscript-domain evaluation and few-shot adaptation. A small subset of crops (24 samples) was excluded from the evaluation because their labels are not present in \ac{snt}.

\section{Evaluation Pipeline and Recognition Strategies}

\begin{figure}[t]
\centering
\includegraphics[width=\textwidth]{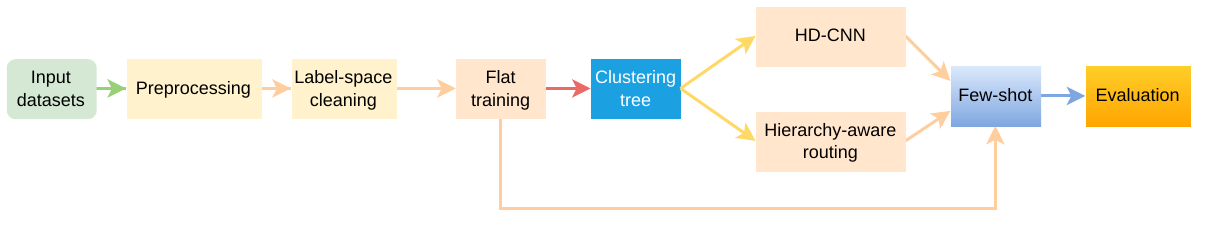}
\caption{Overview of the evaluation pipeline.
Input datasets are preprocessed and cleaned before training the flat model.
The trained flat model provides features for constructing the clustering tree, which is then used by the \ac{hd-cnn} coarse-to-fine model and the hierarchy-aware routing model.
Few-shot adaptation is applied before final evaluation.}
\label{fig:evaluationpipeline}
\end{figure}

This section describes the evaluation pipeline and the three recognition strategies compared in this paper.
The pipeline first standardizes the input images, cleans the \ac{snt} label space, and expands the low-resource \ac{snt} training data through augmentation.
Additionally, flat classifiers are trained as non-hierarchical baselines and also serve as feature extractors for visual clustering. \ac{knn} is evaluated as an additional non-parametric baseline.
The resulting clustering trees are used in two ways: \ac{hd-cnn} uses the leaf and immediate parent levels for coarse-to-fine classification, while the hierarchy-aware routing model uses a deeper tree structure for multi-level local decisions.
\Cref{fig:evaluationpipeline} summarizes the complete evaluation pipeline.

\subsection{Preprocessing and Data Augmentation}
Before model training, the input images from different visual domains are normalized to reduce low-level variation.
For handwritten and manuscript-derived samples, images are converted to grayscale and filtered with a median blur to reduce isolated noise artifacts.
The binarization method is chosen based on the visual properties of each dataset.
For handwritten samples with a relatively uniform background, Otsu thresholding is used~\cite{otsu1979threshold}, whereas for Vergilius Turonensis, where the background is more uneven, and the contrast varies across crops, K-means clustering with two intensity clusters is used to separate foreground strokes from background pixels.

After binarization, each symbol is placed on a square canvas using aspect-ratio-preserving padding and resized to 224 $\times$ 224 pixels.
This keeps the symbol shape from being stretched while providing a common input size for all evaluated backbones.
During training, moderate online transformations, including small rotations, translations, and scaling, are applied to increase variation without intentionally altering the symbol's identity.

In addition to image preprocessing, the \ac{snt} label space is cleaned before training and hierarchy construction.
Some entries in the original inventory are historically or lexically distinct but visually nearly identical after normalization.
Since training an image classifier on such labels can introduce conflicting supervision, a conservative duplicate detection procedure is applied to the clean \ac{snt} standard symbols.
Each image is converted to grayscale, resized to 64 $\times$ 64 pixels, and binarized with a fixed threshold of 128.
Candidate pairs are first filtered by ink ratio, defined as the proportion of black pixels.
Only pairs with an ink-ratio difference of less than 0.02 are compared further.
If their pixel-level similarity is at least 0.998, they are treated as visually equivalent.
For each duplicate group, one representative label is retained, and redundant labels are removed from the training label space.

Additional training samples are generated from the \ac{snt} standard symbols to address the small number of available samples per class.
The augmentation follows the script-level sample generation approach for few-shot handwritten recognition proposed by Chen~\textit{et al.}~\cite{chen2022script}.
For each standard sample, 16 augmented variants are generated.
These samples are used as augmented \ac{snt} training data and are not treated as a separate evaluation domain.

\subsection{Flat Classification Baselines}
\begin{figure}[t]
\centering
\includegraphics[width=0.9\textwidth]{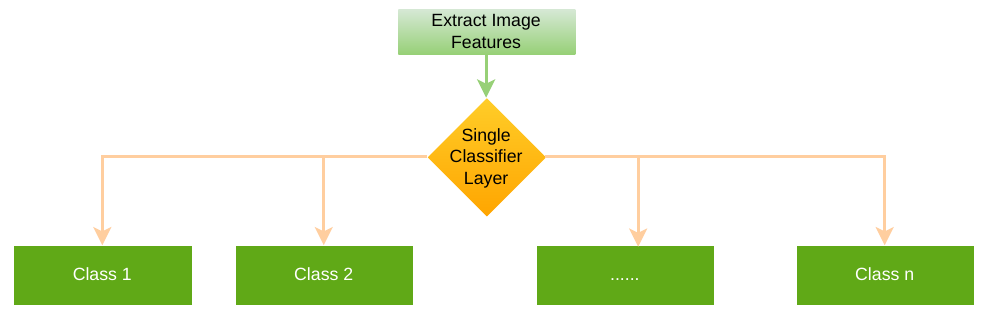}
\caption{Flat classification strategy. After feature extraction, a single global classifier predicts directly over all cleaned leaf classes.}
\label{fig:flatclassification}
\end{figure}

Flat classifiers treat all cleaned leaf labels as independent classes in a single output space.
A visual backbone extracts features from the input image, while a global classification layer directly predicts all target labels.
\cref{fig:flatclassification} illustrates this flat prediction structure.
This setting provides the non-hierarchical baseline for exact Top-k recognition.

The flat baseline is implemented with four lightweight backbone variants: \ac{resnet}18, ConvNeXt-Tiny, \ac{swin}-Tiny, and \ac{vit}-Small. 
They cover different visual families of modern recognition models: residual convolutional networks, modernized convolutional networks, self-attention models, and vision-based transformers~\cite{he2016deep,liu2022convnet,liu2021swin,dosovitskiy2020image}. Their approximate parameter counts are 11.7M, 28.6M, 28.3M, and 22.0M, respectively. All four backbones are initialized from ImageNet-pretrained weights. 
The trained flat models also provide the feature representations used for the subsequent visual clustering step.

All flat models are trained with AdamW~\cite{loshchilov2017decoupled} using a weight decay of $0.05$.
The initial learning rate is set separately for each architecture within the range of $1 \times 10^{-4}$ to $5 \times 10^{-4}$. 
Additionally, a cosine learning-rate schedule with warm-up is used, with a minimum learning rate of $1 \times 10^{-6}$~\cite{loshchilov2016sgdr}.
Training is performed for 50 epochs with automatic mixed precision~\cite{micikevicius2017mixed}.
To handle class imbalance, the cross-entropy loss uses inverse-frequency class weights, and label smoothing with a factor of $0.1$ is applied.

For manuscript-domain adaptation, the trained models are fine-tuned on a small subset of Vergilius Turonensis.
The few-shot protocol is deliberately kept limited to reduce overfitting: Classes with more than five samples provide three fine-tuning samples; classes with exactly five samples provide two samples; classes with two to four samples provide one sample; and singleton classes are kept only for testing.
In total, 117 Vergilius samples from 67 classes are used for fine-tuning and removed from the final evaluation set, leaving 550 test samples across 131 classes. 
The same few-shot protocol is applied to the flat, \ac{hd-cnn}, and hierarchy-aware routing models.

\subsection{Visual Clustering for Hierarchical Grouping}
The \ac{snt} provides a lexicon-based reference structure for Tironian notes and is retained as an external reference for group-level evaluation. However, the hierarchical groupings used by the recognition models are generated from visual feature similarity. This keeps the grouping aligned with the image recognition task.

A trained flat model is used as a feature extractor by removing its final classification head.
For each retained \ac{snt} class, a feature vector is extracted from the representative standard symbol image.
The feature vectors $\bm{x}$ are L2-normalized ($\hat{\bm{x}}$), and pairwise cosine distances are computed as

\begin{equation}
D_{ij} = 1 - \hat{\bm{x}}_i^\top \hat{\bm{x}}_j .
\end{equation}

The resulting distance matrix $D$ is used for agglomerative clustering with average linkage~\cite{mullner2011modern,rokach2005clustering}.
Average linkage defines the distance between two clusters as the average pairwise distances between their members, providing a balanced grouping criterion for visually related classes.

The clustering tree is generated recursively.
For the \ac{resnet}18-based tree, a distance-threshold strategy is used, with the threshold adjusted across hierarchy depth.
For ConvNeXt, \ac{swin}, and \ac{vit} features, the same threshold-based strategy produced excessively deep and unbalanced trees. A dynamic $K$ strategy is therefore used to control the local branching factor.
For a node containing $N$ classes, the target number of child clusters is set to $K=\bigl\lceil\sqrt{N}\,\bigr\rceil$.
For implementation, each resulting clustering tree is stored as a nested structure and provides the visual label organization used by both non-flat recognition strategies.

\begin{figure}[t]
\centering
\includegraphics[width=\textwidth]{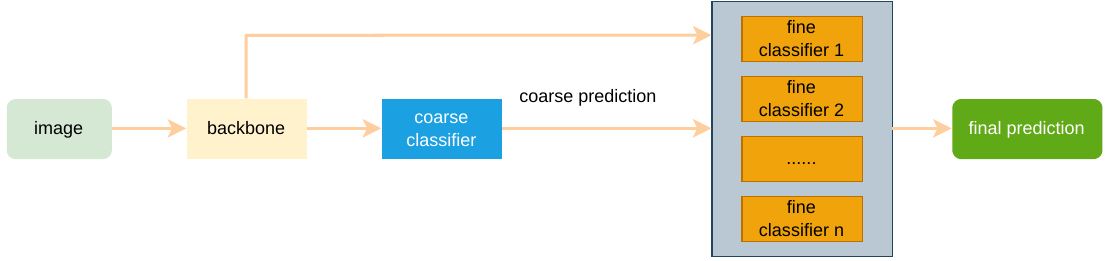}
\caption{\ac{hd-cnn}-style coarse-to-fine classification strategy. A shared backbone feeds a coarse classifier that predicts the parent group, while group-specific specialist classifiers predict among the corresponding fine classes.}
\label{fig:hdcnnclassification}
\end{figure}
\subsection{\ac{hd-cnn} Coarse-to-Fine Classification}
\ac{hd-cnn} is used as a two-level coarse-to-fine recognition strategy~\cite{yan2015hd}.
In this setting, the two-level label structure is derived from the generated visual clustering tree.
The leaf labels define the fine classes, while the immediate parent nodes of these leaves define the coarse symbol groups.
Each fine class is therefore mapped to one coarse group.

The model consists of a shared visual backbone, one coarse classifier, and several fine-level specialist classifiers.
\Cref{fig:hdcnnclassification} shows the \ac{hd-cnn}-style coarse-to-fine structure used in this paper.
During training, the model receives both the coarse group label and the fine leaf label.
The coarse classifier predicts the parent group of the input symbol, while each specialist classifier distinguishes only the fine classes assigned to one parent group.

During inference, each leaf class receives a global score by combining the probability of its coarse group with the probability assigned by the corresponding fine specialist.
This produces a ranked prediction list across all fine classes, so the model can be evaluated using the same Top-1, Top-5, Top-10, and Top-30 metrics as the flat classifiers.
Compared with deeper routing, this two-level structure uses hierarchical information while limiting the number of sequential decisions.

\subsection{Hierarchy-Aware Routing Classification}
\begin{figure}[t]
\centering
\includegraphics[width=\textwidth]{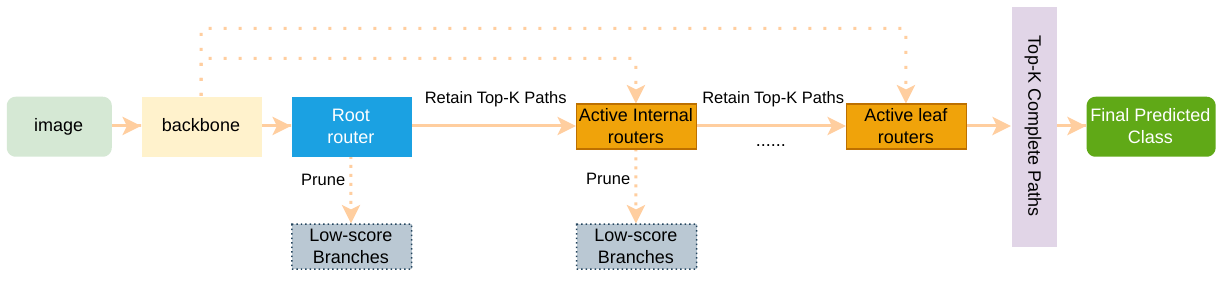}
\caption{Hierarchy-aware routing classification strategy. The model follows local routing decisions through a deeper clustering tree and retains top-ranked candidate paths during beam-search inference.}
\label{fig:hierarchyrouting}
\end{figure}

The hierarchy-aware routing model uses a deeper part of the generated clustering tree.
Instead of using a single global output layer to predict all classes directly, the model replaces the flat classifier with node-specific routing classifiers.
Each internal node of the tree has its own classifier, and each classifier predicts among the child nodes of that internal node.
A prediction is obtained by following a sequence of local routing decisions from the root of the tree towards a leaf class.
\Cref{fig:hierarchyrouting} illustrates this hierarchy-aware routing strategy.

During training, each ground-truth label is converted into its target path in the clustering tree.
Only the routing classifiers along this target path contribute to the loss.
Since the generated trees are not necessarily balanced, different leaf classes may have target paths of different lengths.
A simple sum of local losses would give deeper paths a larger contribution.
To reduce this depth-related imbalance, the model uses a path-normalized hierarchical loss:

\begin{equation}
L_i^\text{hier} = \frac{1}{|P_i|}\sum_{p \in P_i} \ell_{i,p}\;,
\end{equation}
where $P_i$ is the target path of sample $i$ and $\ell_{i,p}$ is the local cross-entropy loss at node $p$.
The final batch loss is obtained by averaging the normalized sample losses.
Node-specific class weights can also be used inside local routing classifiers when the child branches of a node are imbalanced.

During inference, greedy prediction and beam search are used in different phases.
Greedy inference follows the locally highest-scoring child at each node, while beam search retains several candidate paths and therefore provides a more robust estimate of the final leaf prediction.
The hierarchy-aware routing model is evaluated at the leaf level with the same Top-k metrics as the other strategies.
In addition, group-level scores are reported for the routing model to examine whether incorrect leaf predictions still remain within related visual groups derived from the reference hierarchy.

\section{Results and Discussion}
This section reports Top-k accuracies for flat classification, \ac{hd-cnn} coarse-to-fine classification, and hierarchy-aware routing classification.
Since handwritten performance is near-perfect in most, the discussion focuses on Vergilius Turonensis, which represents the strongest domain shift.
All values are Top-k accuracies in percent. Results are averaged over four runs, except for \ac{hd-cnn}, which was averaged over three runs.

For the hierarchy-aware experiments, results are reported at the leaf level and, where applicable, at the group level.
Cluster-group accuracy is computed using the visual clustering tree generated during training.
\ac{snt}-group accuracy uses the structured representation of the \ac{snt} reference hierarchy as an external comparison.
This metric is interpreted cautiously, because repeated or overlapping parent assignments make it more permissive than the generated cluster-group metric.

\subsection{Flat Classification}
The flat classifiers provide the non-hierarchical baseline.
Tables~\ref{tab:flat-without-fewshot} and~\ref{tab:flat-with-fewshot} show the results without, and with few-shot fine-tuning.

\begin{table}[tb]
\caption{Flat classification performance without few-shot fine-tuning. }
    \label{tab:flat-without-fewshot}
    \centering
    \begin{minipage}{0.48\textwidth}
\centering
\resizebox{\linewidth}{!}{%
\begin{tabular}{lcccc}
\toprule
\multicolumn{5}{c}{\textbf{Vergilius Turonensis}} \\ %
\midrule
Model & Top-1 & Top-5 & Top-10 & Top-30 \\
\midrule
\ac{resnet}18 & 36.28 & 46.67 & 50.45 & 56.30 \\
ConvNeXt & 40.37 & 50.75 & 54.87 & 62.93 \\
\ac{swin} & \textbf{43.82}& 55.17 & 60.65 & \textbf{69.83}\\
\ac{vit} & 42.62 & 52.81 & 59.15 & 67.70 \\
\bottomrule
\end{tabular}
}
\end{minipage}%
\hspace{0.1cm}%
\begin{minipage}{0.48\textwidth}
\centering
\resizebox{\linewidth}{!}{%
\begin{tabular}{lcccc}
\toprule
\multicolumn{5}{c}{\textbf{Handwritten}} \\
\midrule
Model & Top-1 & Top-5 & Top-10 & Top-30 \\
\midrule
\ac{resnet}18 & 98.99 & 99.91 & 99.94 & 99.96 \\
ConvNeXt & 99.36 & 99.88 & 99.91 & 99.93 \\
\ac{swin} & 99.03 & 99.91 & 99.92 & 99.93 \\
\ac{vit} & 99.27 & 99.94 & 99.95 & 99.96 \\
\bottomrule
\end{tabular}
}
\end{minipage}
\end{table}
Without few-shot fine-tuning, the flat models achieve very high accuracy on the handwritten test set, with Top-1 between 98.99\,\% and 99.36\,\%.
In contrast, Top-1 accuracy on Vergilius Turonensis ranges only between 36.28\,\% and 43.82\,\%.
This confirms that clean handwritten recognition and historical manuscript recognition behave differently.
The manuscript crops contain uneven background texture, lower contrast, degraded strokes, incomplete local details, and stronger ambiguity between visually similar symbols.

Few-shot fine-tuning substantially improves the manuscript-domain recognition for all flat models.
On Vergilius Turonensis, Top-1 accuracy increases from 36.28\,\% to 68.73\,\% for \ac{resnet}18, from 40.37\,\% to 79.14\,\% for ConvNeXt, from 43.82\,\% to 82.09\,\% for \ac{swin}, and from 42.62\,\% to 81.77\,\% for \ac{vit}.
The handwritten results remain close to the maximum value, so the main effect of target-domain transfer is visible on Vergilius Turonensis.
These results show that even limited manuscript-domain supervision helps the models adapt to manuscript-specific background, stroke, and contrast conditions.
\begin{table}[tb]
\caption{Flat classification performance after few-shot fine-tuning. }
\label{tab:flat-with-fewshot}
\centering
\small
\begin{minipage}{0.48\textwidth}
\centering
\resizebox{\linewidth}{!}{%
\begin{tabular}{lcccc}
\toprule
\multicolumn{5}{c}{\textbf{Vergilius Turonensis}} \\
\midrule
Model & Top-1 & Top-5 & Top-10 & Top-30 \\
\midrule
\ac{resnet}18 & 68.73 & 82.36 & 85.68 & 89.14 \\
ConvNeXt & 79.14 & 86.14 & 87.32 & 89.14 \\
\ac{swin} & \textbf{82.09}& 87.59 & 89.36 & \textbf{90.59}\\
\ac{vit} & 81.77 & 88.09 & 88.96 & 90.55 \\
\bottomrule
\end{tabular}
}
\end{minipage}
\hspace{0.1cm}%
\begin{minipage}{0.48\textwidth}
\centering
\resizebox{\linewidth}{!}{%
\begin{tabular}{lcccc}
\toprule
\multicolumn{5}{c}{\textbf{Handwritten}} \\
\midrule
Model & Top-1 & Top-5 & Top-10 & Top-30 \\
\midrule
\ac{resnet}18 & 97.98 & 99.84 & 99.91 & 99.96 \\
ConvNeXt & 99.34 & 99.89 & 99.92 & 99.96 \\
\ac{swin} & 99.03 & 99.92 & 99.95 & 99.96 \\
\ac{vit} & 99.32 & 99.94 & 99.97 & 99.97 \\
\bottomrule
\end{tabular}
}
\end{minipage}
\end{table}

Pixel-level \ac{knn} reaches 90.12\% accuracy on the handwritten data but only 4.20\% on the Vergilius data. Feature-based \ac{knn} improves performance on Vergilius to 38.83\%, but its performance remains limited. 

To account for the class imbalance in the evaluation sets, balanced accuracy is additionally considered. For the best flat model on Vergilius, \ac{swin} achieves 43.82\% Top-1 accuracy without few-shot adaptation, while its balanced accuracy is 29.37\%. After few-shot adaptation, \ac{swin} reaches 82.09\% Top-1 accuracy and 51.13\% balanced accuracy. The lower balanced scores indicate that rare classes remain more challenging, although few-shot adaptation still improves performance across the class distribution. 

\subsection{Hierarchy-Aware Routing Classification}

The hierarchy-aware routing experiments compare self-generated trees, a shared tree from ImageNet-pretrained \ac{resnet}18 features \cite{deng2009imagenet,he2016deep}, and a shared tree from the trained \ac{resnet}18 flat classifier.
The self-generated setting allows each backbone to use its own feature space, while the shared-tree settings keep the hierarchy fixed across backbones.

This shared-tree setting is used to keep the hierarchy fixed across backbones, so that the comparison is not confounded by independently generated tree structures. \ac{resnet}18 is chosen as the reference because it serves as the lightweight convolutional baseline and provides a simple common feature space for constructing the shared hierarchy. This choice is not based on its flat classification accuracy and does not imply that the \ac{resnet}18-derived tree is globally optimal. The pretrained and trained \ac{resnet}18 trees further allow us to compare generic and task-adapted visual features.

\begin{figure}[t]
\centering
\includegraphics[width=0.92\textwidth]{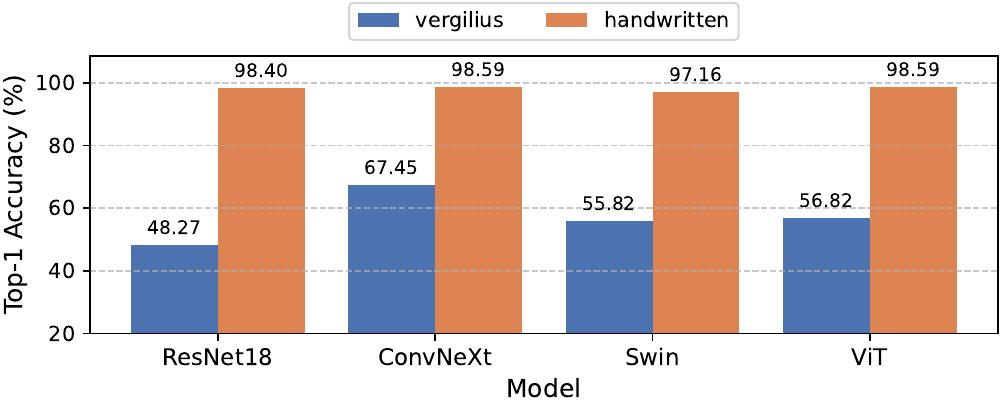}
\caption{Leaf-level Top-1 accuracy of hierarchy-aware models with self-generated clustering trees after few-shot fine-tuning.}
\label{fig:self_generated_fewshot}
\end{figure}

\Cref{fig:self_generated_fewshot} summarizes the self-generated tree setting after few-shot fine-tuning.
The handwritten results remain high for all models, whereas the Vergilius Turonensis results show clearer differences between the backbones.
In this setting, ConvNeXt obtains the highest Top-1 accuracy on Vergilius Turonensis.

The main shared-tree setting uses the clustering tree generated from the trained \ac{resnet}18 flat model.
\Cref{tab:baseline-tree-compact} summarizes the Vergilius Turonensis results without, and with few-shot fine-tuning.
To keep the table compact, only Top-1 and Top-30 values are shown.

\begin{table}[tb]
\caption{Hierarchy-aware routing performance on Vergilius Turonensis using the shared tree generated from the trained \ac{resnet}18 flat model.
}
\label{tab:baseline-tree-compact}
\centering
\setlength{\tabcolsep}{3pt}
\begin{tabular*}{\textwidth}{@{\extracolsep{\fill}}llcccc@{}}
\toprule
 & & \multicolumn{2}{l}{\textbf{Without few-shot}} & \multicolumn{2}{l}{\textbf{With few-shot}} \\
\cmidrule{3-4}\cmidrule{5-6}
Model & Level & Top-1 & Top-30 & Top-1 & Top-30 \\
\midrule
\ac{resnet}18 & Leaf & 37.59 & 56.45 & 48.27 & 70.82 \\
 & Cluster group & 44.76 & 65.41 & 55.68 & 76.27 \\
 & \ac{snt} group & 44.14 & 77.56 & 53.55 & 82.65 \\
ConvNeXt & Leaf & 36.99 & 54.34 & \textbf{71.82} & \textbf{85.05} \\
 & Cluster group & 40.59 & 57.91 & 74.18 & 85.95 \\
 & \ac{snt} group & 46.89 & 72.74 & 75.91 & 89.07 \\
\ac{swin} & Leaf & \textbf{43.78} & \textbf{62.07} & 69.91 & 83.04 \\
 & Cluster group & 47.79 & 65.29 & 71.00 & 84.45 \\
 & \ac{snt} group & 54.73 & 79.92 & 74.41 & 89.25 \\
\ac{vit} & Leaf & 37.93 & 57.98 & 67.68 & 80.09 \\
 & Cluster group & 47.60 & 61.84 & 69.00 & 81.82 \\
 & \ac{snt} group & 45.80 & 72.93 & 71.54 & 86.52 \\
\bottomrule
\end{tabular*}
\end{table}

Few-shot fine-tuning improves hierarchy-aware results across all four backbones, but it does not consistently outperform flat classification at the leaf level.
Group-level scores are usually higher than the corresponding leaf-level scores, indicating that some errors remain within visually or structurally related groups.
One possible reason is error propagation: if an early routing decision selects an incorrect branch, the final prediction is restricted to that part of the tree.

\begin{figure}[htbp]
\centering
\includegraphics[width=0.9\textwidth]{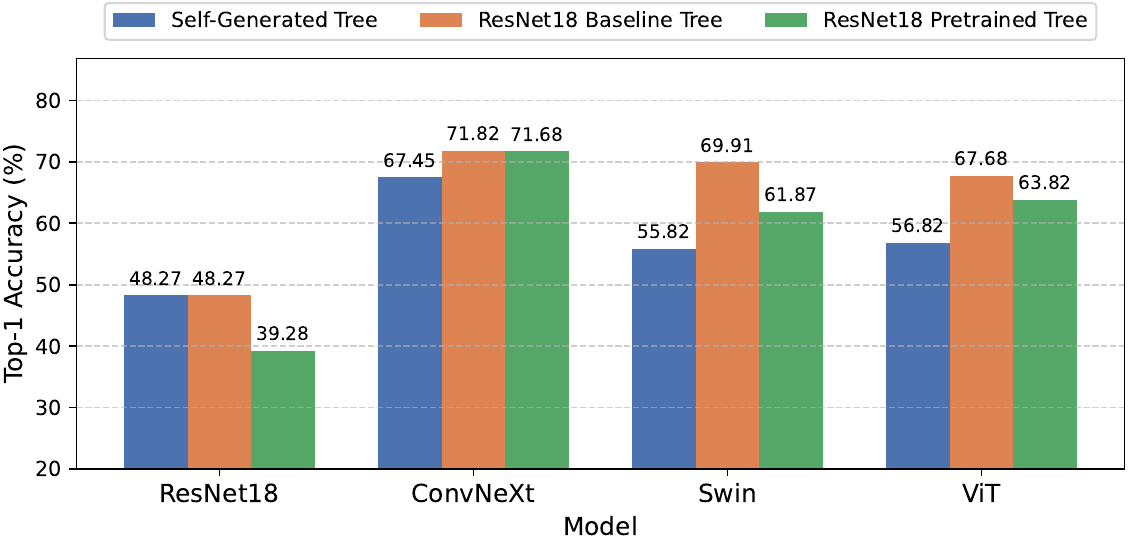}
\caption{Comparison of different clustering tree sources on Vergilius Turonensis after few-shot fine-tuning.}
\label{fig:tree_comparison}
\end{figure}

\Cref{fig:tree_comparison} compares the three tree sources after few-shot fine-tuning.
For ConvNeXt, \ac{swin}, and \ac{vit}, the shared tree generated from the trained \ac{resnet}18 flat model performs better than that generated from \ac{resnet}18 pretrained features.
This suggests that task-adapted visual features provide a more suitable hierarchy than generic pretrained features for the Tironian notes recognition task.

\subsection{\ac{hd-cnn} Coarse-to-Fine Classification}
\Cref{tab:hdcnn-vergilius} shows the \ac{hd-cnn} results on Vergilius Turonensis. The models use a two-level coarse-to-fine structure derived from the lowest levels of the generated clustering tree.

\begin{table}[tb]
\caption{\ac{hd-cnn} coarse-to-fine classification performance on Vergilius Turonensis.}
\label{tab:hdcnn-vergilius}
\centering
\small
\begin{tabular*}{\textwidth}{@{\extracolsep{\fill}}lcccccccc@{}}
\toprule
& \multicolumn{4}{c}{\textbf{Without few-shot}} & \multicolumn{4}{c}{\textbf{With few-shot}} \\
\cmidrule{2-5}\cmidrule{6-9}
Model & Top-1 & Top-5 & Top-10 & Top-30 & Top-1 & Top-5 & Top-10 & Top-30 \\
\midrule
\ac{resnet}18 & 37.68 & 46.78 & 50.13 & 57.52 & 67.21 & 76.91 & 80.91 & 86.43 \\
ConvNeXt & 36.53 & 45.53 & 47.88 & 52.27 & 66.43 & 78.55 & 81.45 & 86.12 \\
\ac{swin} & \textbf{45.43} & 56.82 & 60.47 & \textbf{67.37} & \textbf{75.76} & 82.24 & 85.15 & \textbf{87.94} \\
\ac{vit} & 43.68 & 56.27 & 59.62 & 63.57 & 74.24 & 80.49 & 82.49 & 86.43\\
\bottomrule
\end{tabular*}
\end{table}

Without few-shot adaptation, \ac{hd-cnn} compares favorably with flat classification on Vergilius Turonensis.
It improves Top-1 accuracy for \ac{resnet}18, \ac{swin}, and \ac{vit}, with the highest performance achieved by \ac{swin} at 45.43\,\%.
ConvNeXt is the exception in which the flat classifier performs better.
However, after few-shot fine-tuning, the flat models show stronger performance in the manuscript domain than \ac{hd-cnn}.
This indicates that the relative advantage of coarse-to-fine classification depends on the adaptation setting.

The comparison with hierarchy-aware routing suggests that the shallow two-level \ac{hd-cnn} structure is more stable than deeper routing in the reported results on the manuscript domain.
\ac{hd-cnn} uses the generated tree only to define coarse groups and fine classes, while the routing model follows several local decisions through a deeper tree, which reduces the number of sequential decisions in \ac{hd-cnn} and limits the effect of early routing errors.

\begin{figure}[t]
\centering
\includegraphics[width=0.92\textwidth]{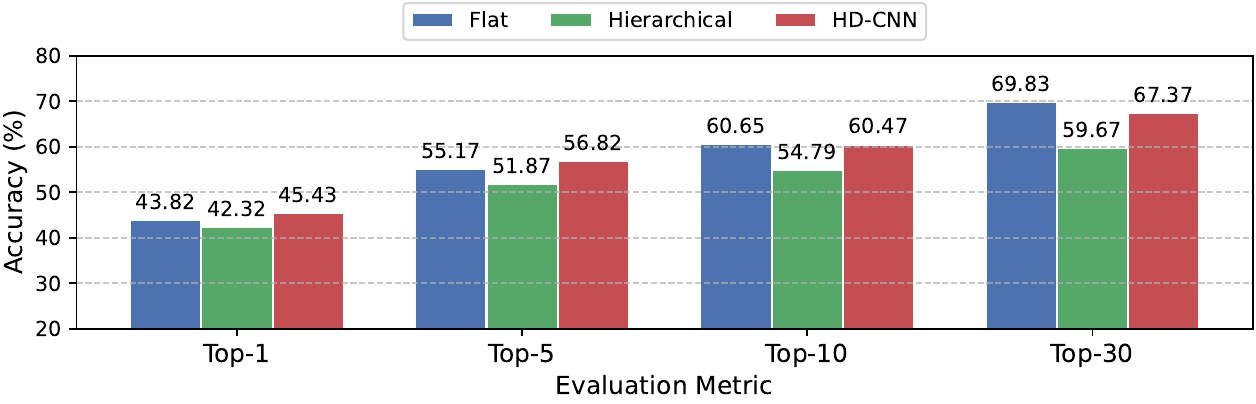}
\caption{Top-k accuracy of \ac{swin}-based models on Vergilius Turonensis without few-shot.}
\label{fig:vergilius_hdcnn_topk}
\end{figure}

\begin{table}[htbp]
\caption{Representative misclassified examples, including manuscript crops from Vergilius Turonensis~\cite{ecodices_vergilius} and corresponding standard symbols from \ac{snt}~\cite{hellmann_martinellus}.}
\label{tab:misclassified_examples}

\centering
\small
\setlength{\tabcolsep}{4pt}

\begin{tabular}{lcccccc}
\toprule
\raisebox{0.5\height}{Label} &
\raisebox{0.5\height}{Image} &
\shortstack{Flat\\Predict} &
\shortstack{Hierarchical\\Predict} &
\raisebox{0.5\height}{True} &
\shortstack{True-label\\Rank (Flat)} &
\shortstack{True-label\\Rank (Hier.)} \\
\midrule

sit &
\raisebox{-0.3\height}{\includegraphics[width=0.6cm]{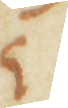}}&
\raisebox{-0.3\height}{\includegraphics[width=0.6cm]{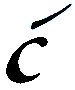}}&
\raisebox{-0.3\height}{\includegraphics[width=0.5cm]{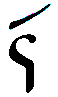}} &
\raisebox{-0.3\height}{\includegraphics[width=0.5cm]{bf_sit}} &
4 &
1 \\

deus &
\raisebox{-0.3\height}{\includegraphics[width=1.2cm]{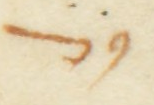}}&
\raisebox{-0.3\height}{\includegraphics[width=1.0cm]{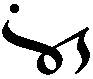}}&
\raisebox{-0.3\height}{\includegraphics[width=1.0cm]{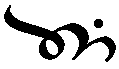}}&
\raisebox{-0.3\height}{\includegraphics[width=1.0cm]{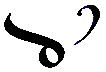}}&
13 &
3\\

templum &
\raisebox{-0.3\height}{\includegraphics[width=1.5cm]{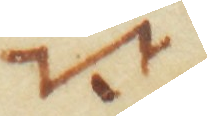}}&
\raisebox{-0.3\height}{\includegraphics[width=1.0cm]{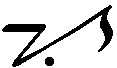}}&
\raisebox{-0.3\height}{\includegraphics[width=1.0cm]{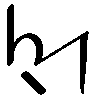}}&
\raisebox{-0.3\height}{\includegraphics[width=1.0cm]{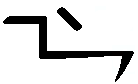}}&
$>30$ &
$>30$ \\

\bottomrule
\end{tabular}
\end{table}

\Cref{fig:vergilius_hdcnn_topk} provides a \ac{swin}-based Top-k comparison between the three recognition strategies before few-shot adaptation. \ac{hd-cnn} achieves the highest Top-1 and Top-5 accuracies in this comparison, while the flat \ac{swin} classifier obtains slightly higher Top-10 and Top-30 results. This suggests that the coarse-to-fine structure can improve the highest-ranked predictions, although the flat model still retains stronger broader candidate coverage at larger Top-k values.

\subsection{Error Analysis}
Representative challenging examples are shown in Table~\ref{tab:misclassified_examples}.
Each row presents a manuscript sample, the flat and hierarchy-aware predictions, the standard symbol and the true-label rank. The examples illustrate visually ambiguity in the manuscript domain: some crops resemble symbols with different labels or differ strongly from the standard symbols. In the first two examples, the hierarchy-aware models ranks the true label higher than flat models. This may because hierarchical models narrow the prediction to a visually related subgroup, which can help keep similar candidates closer together. In the last example the true label is outside the Top-30 candidates for both models. This suggests that some errors are caused by ambiguous manuscript evidence.

\section{Conclusion and Outlook}
This paper investigated Tironian note recognition as a low-resource visual classification problem using clean \ac{snt} standard symbols, augmented \ac{snt} samples, modern handwritten notes, and manuscript crops from Vergilius Turonensis.
The study compared flat classification, \ac{hd-cnn}-style coarse-to-fine classification, and hierarchy-aware routing classification under comparable training and few-shot adaptation conditions.

The results show that the visual domain has a strong influence on recognition performance, since all strategies achieve high accuracy on the cleaner handwritten test set, while the recognition on the Vergilius Turonensis remains substantially more difficult because of background texture and degraded strokes.
Without few-shot adaptation, \ac{hd-cnn} performs competitively and achieves the best Top-1 accuracy.
However, after few-shot adaptation, flat models achieve the strongest performance in the manuscript domain, showing that even limited target-domain supervision is highly effective.

The hierarchy-aware routing model does not consistently improve final leaf-level recognition in the current setting, which suggests that deeper routing can introduce additional instability when early decisions send samples into incorrect branches.
Nevertheless, the group-level results are useful for analyzing whether errors remain within visually or structurally related symbols, which is important for Tironian notes.

The main limitations are the small amount of annotated manuscript-domain samples, the focus on a single manuscript source, and the dependence of non-flat models on the quality of the visual clustering tree.
Future work should, therefore, evaluate the approaches on additional manuscripts, increase coverage for rare symbols, and investigate hierarchy construction methods that combine learned visual similarity with the \ac{snt} reference structure and historical knowledge.

Overall, the results indicate that hierarchical structure can support Tironian note recognition, especially for error analysis and non-adapted recognition.
Its final accuracy depends on tree construction, routing depth, backbone architecture, and the amount of available supervision in the manuscript domain.

\section*{Acknowledgements}
The authors gratefully acknowledge the scientific support and \textsc{hpc} resources provided by the Erlangen National High Performance Computing Center (\textsc{nhr@fau}) of the Friedrich-Alexander-Universität Erlangen-Nürnberg (\textsc{fau}).
The hardware is funded by the German Research Foundation (\textsc{dfg}).
This work was partially funded by VolkswagenStiftung as part of the project ``Stenographie in historischen Dokumenten. Entwicklung eines Kurzschrifttools auf Grundlage der Dechiffrierung eines Vergilkommentars in tironischen Noten'' (funding number: A141503).
We thank Martin Hellmann for providing access to \acl{snt}.

 \bibliographystyle{plain}
 \bibliography{mybibliography}

\begin{thebibliography}{10}

\bibitem{chen2022script}
Wei Chen, Xiangdong Su, and Haoran Zhang.
\newblock Script-level word sample augmentation for few-shot handwritten text
  recognition.
\newblock In {\em International Conference on Frontiers in Handwriting
  Recognition}, pages 316--330. Springer, 2022.

\bibitem{deng2009imagenet}
Jia Deng, Wei Dong, Richard Socher, Li-Jia Li, Kai Li, and Li~Fei-Fei.
\newblock Imagenet: A large-scale hierarchical image database.
\newblock In {\em 2009 IEEE conference on computer vision and pattern
  recognition}, pages 248--255. Ieee, 2009.

\bibitem{deng2011fast}
Jia Deng, Sanjeev Satheesh, Alexander Berg, and Fei Li.
\newblock Fast and balanced: Efficient label tree learning for large scale
  object recognition.
\newblock {\em Advances in neural information processing systems}, 24, 2011.

\bibitem{dosovitskiy2020image}
Alexey Dosovitskiy, Lucas Beyer, Alexander Kolesnikov, Dirk Weissenborn,
  Xiaohua Zhai, Thomas Unterthiner, Mostafa Dehghani, Matthias Minderer, Georg
  Heigold, Sylvain Gelly, et~al.
\newblock An image is worth 16x16 words: Transformers for image recognition at
  scale.
\newblock {\em arXiv preprint arXiv:2010.11929}, 2020.

\bibitem{ecodices_vergilius}
{e-codices}.
\newblock Bern, burgerbibliothek, cod. 165: Vergilius: Bucolica, georgica,
  aeneis / scholia turonensia.
\newblock Online manuscript facsimile, 2015.
\newblock Online since Dec. 17, 2015. Accessed: Apr. 30, 2026.

\bibitem{he2016deep}
Kaiming He, Xiangyu Zhang, Shaoqing Ren, and Jian Sun.
\newblock Deep residual learning for image recognition.
\newblock In {\em IEEE conference on computer vision and pattern recognition},
  pages 770--778, 2016.

\bibitem{hellmann_martinellus}
Martin Hellmann.
\newblock Hypertext-lexikon der tironischen noten.
\newblock [Online]. Available: \url{https://martinellus.de/snt2/n/incipit.htm}.
\newblock Accessed: Apr. 27, 2026.

\bibitem{khandagale2020bonsai}
Sujay Khandagale, Han Xiao, and Rohit Babbar.
\newblock Bonsai: diverse and shallow trees for extreme multi-label
  classification.
\newblock {\em Machine Learning}, 109(11):2099--2119, 2020.

\bibitem{lecun2015deep}
Yann LeCun, Yoshua Bengio, and Geoffrey Hinton.
\newblock Deep learning.
\newblock {\em nature}, 521(7553):436--444, 2015.

\bibitem{lecun2002gradient}
Yann LeCun, L{\'e}on Bottou, Yoshua Bengio, and Patrick Haffner.
\newblock Gradient-based learning applied to document recognition.
\newblock {\em IEEE}, 86(11):2278--2324, 2002.

\bibitem{liu2013probabilistic}
Baoyuan Liu, Fereshteh Sadeghi, Marshall Tappen, Ohad Shamir, and Ce~Liu.
\newblock Probabilistic label trees for efficient large scale image
  classification.
\newblock In {\em IEEE conference on computer vision and pattern recognition},
  pages 843--850, 2013.

\bibitem{liu2021swin}
Ze~Liu, Yutong Lin, Yue Cao, Han Hu, Yixuan Wei, Zheng Zhang, Stephen Lin, and
  Baining Guo.
\newblock Swin transformer: Hierarchical vision transformer using shifted
  windows.
\newblock In {\em IEEE/CVF international conference on computer vision}, pages
  10012--10022, 2021.

\bibitem{liu2022convnet}
Zhuang Liu, Hanzi Mao, Chao-Yuan Wu, Christoph Feichtenhofer, Trevor Darrell,
  and Saining Xie.
\newblock A convnet for the 2020s.
\newblock In {\em IEEE/CVF conference on computer vision and pattern
  recognition}, pages 11976--11986, 2022.

\bibitem{lombardi2020deep}
Francesco Lombardi and Simone Marinai.
\newblock Deep learning for historical document analysis and recognition—a
  survey.
\newblock {\em Journal of Imaging}, 6(10):110, 2020.

\bibitem{loshchilov2016sgdr}
Ilya Loshchilov and Frank Hutter.
\newblock Sgdr: Stochastic gradient descent with warm restarts.
\newblock {\em arXiv preprint arXiv:1608.03983}, 2016.

\bibitem{loshchilov2017decoupled}
Ilya Loshchilov and Frank Hutter.
\newblock Decoupled weight decay regularization.
\newblock {\em arXiv preprint arXiv:1711.05101}, 2017.

\bibitem{micikevicius2017mixed}
Paulius Micikevicius, Sharan Narang, Jonah Alben, Gregory Diamos, Erich Elsen,
  David Garcia, Boris Ginsburg, Michael Houston, Oleksii Kuchaiev, Ganesh
  Venkatesh, et~al.
\newblock Mixed precision training.
\newblock {\em arXiv preprint arXiv:1710.03740}, 2017.

\bibitem{mullner2011modern}
Daniel M{\"u}llner.
\newblock Modern hierarchical, agglomerative clustering algorithms.
\newblock {\em arXiv preprint arXiv:1109.2378}, 2011.

\bibitem{otsu1979threshold}
Nobuyuki Otsu et~al.
\newblock A threshold selection method from gray-level histograms.
\newblock {\em Automatica}, 11(285-296), 1979.

\bibitem{pearse_cnt}
R.~Pearse.
\newblock An ancient handbook of short-hand: Tironian notes and the commentarii
  notarum tironianarum.
\newblock [Online]. Available:
  \url{https://www.roger-pearse.com/weblog/2019/08/14/an-ancient-handbook-of-short-hand-tironian-notes-and-the-commentarii-notarum-tironianarum/},
  August 2019.
\newblock Accessed: Apr. 27, 2026.

\bibitem{prabhu2018parabel}
Yashoteja Prabhu, Anil Kag, Shrutendra Harsola, Rahul Agrawal, and Manik Varma.
\newblock Parabel: Partitioned label trees for extreme classification with
  application to dynamic search advertising.
\newblock In {\em 2018 World Wide Web Conference}, pages 993--1002, 2018.

\bibitem{rokach2005clustering}
Lior Rokach and Oded Maimon.
\newblock Clustering methods.
\newblock In {\em Data mining and knowledge discovery handbook}, pages
  321--352. Springer, 2005.

\bibitem{roy2020tree}
Deboleena Roy, Priyadarshini Panda, and Kaushik Roy.
\newblock Tree-cnn: a hierarchical deep convolutional neural network for
  incremental learning.
\newblock {\em Neural networks}, 121:148--160, 2020.

\bibitem{schmitz1893commentarii}
Wilhelm Schmitz, editor.
\newblock {\em Commentarii notarum Tironianarum cum prolegomenis,
  adnotationibus criticis et exegeticis, notarumque indice alphabetico}.
\newblock B. G. Teubner, Lipsiae, 1893.
\newblock Digitized by Boston Public Library. Accessed: Apr. 29, 2026.

\bibitem{shorten2019survey}
Connor Shorten and Taghi~M Khoshgoftaar.
\newblock A survey on image data augmentation for deep learning.
\newblock {\em Journal of big data}, 6(1):1--48, 2019.

\bibitem{silla2011survey}
Carlos~N Silla~Jr and Alex~A Freitas.
\newblock A survey of hierarchical classification across different application
  domains.
\newblock {\em Data mining and knowledge discovery}, 22(1):31--72, 2011.

\bibitem{sun2013find}
Min Sun, Wan Huang, and Silvio Savarese.
\newblock Find the best path: An efficient and accurate classifier for image
  hierarchies.
\newblock In {\em IEEE International Conference on Computer Vision}, pages
  265--272, 2013.

\bibitem{yan2015hd}
Zhicheng Yan, Hao Zhang, Robinson Piramuthu, Vignesh Jagadeesh, Dennis DeCoste,
  Wei Di, and Yizhou Yu.
\newblock Hd-cnn: hierarchical deep convolutional neural networks for large
  scale visual recognition.
\newblock In {\em IEEE international conference on computer vision}, pages
  2740--2748, 2015.

\bibitem{zhu2017b}
Xinqi Zhu and Michael Bain.
\newblock B-cnn: branch convolutional neural network for hierarchical
  classification.
\newblock {\em arXiv preprint arXiv:1709.09890}, 2017.

\end{thebibliography}

\end{document}

\maketitle
\section{Supplementary Results}

The generated visual clustering trees differ substantially from the SNT reference tree. The SNT tree follows the organization provided by the source glossary, whereas the generated trees are constructed from visual feature similarities and are used as recognition-oriented hierarchies to guide the routing decisions of the hierarchy-aware models. Table~\ref{tab:tree_structure_comparison} compares the number of nodes at each hierarchy level. The SNT tree expands strongly at the lower levels, especially at Levels 3 and 4, while the generated clustering trees remain much more compact.
\begin{table}[htbp]
\caption{Structural comparison of the SNT reference tree and generated visual clustering trees.}
\label{tab:tree_structure_comparison}
\centering
\small
\begin{tabular}{lrrrr}
\toprule
Tree type & Level 1 & Level 2 & Level 3 & Level 4 \\
\midrule
SNT tree& 26 & 392 & 4,968 & 10,678 \\
ResNet18 clustering tree & 108 & 355 & 576 & 182 \\
ConvNeXt clustering tree & 40 & 612 & 537 & 58 \\
Swin clustering tree & 40 & 556 & 596 & 7 \\
ViT clustering tree & 40 & 655 & 521 & 8 \\
\bottomrule
\end{tabular}
\end{table}

\begin{table}[htbp]
\caption{Flat classification performance without few-shot fine-tuning.}
\label{tab:supp_flat_without_fewshot}
\centering
\small
\begin{tabular}{lcccc}
\toprule
\multicolumn{5}{c}{Vergilius Turonensis} \\
\midrule
Model & Top-1 & Top-5 & Top-10 & Top-30 \\
\midrule
ResNet18 & 36.28 $\pm$ 0.74 & 46.67 $\pm$ 1.02 & 50.45 $\pm$ 0.77 & 56.30 $\pm$ 1.13 \\
ConvNeXt & 40.37 $\pm$ 0.97 & 50.75 $\pm$ 1.53 & 54.87 $\pm$ 1.01 & 62.93 $\pm$ 1.21 \\
Swin & 43.82 $\pm$ 0.65 & 55.17 $\pm$ 0.51 & 60.65 $\pm$ 0.40 & 69.83 $\pm$ 1.64 \\
ViT & 42.62 $\pm$ 0.93 & 52.81 $\pm$ 0.80 & 59.15 $\pm$ 0.34 & 67.70 $\pm$ 0.62 \\
\midrule
\multicolumn{5}{c}{Handwritten} \\
\midrule
Model & Top-1 & Top-5 & Top-10 & Top-30 \\
\midrule
ResNet18 & 98.99 $\pm$ 0.10 & 99.91 $\pm$ 0.02 & 99.94 $\pm$ 0.00 & 99.96 $\pm$ 0.00 \\
ConvNeXt & 99.36 $\pm$ 0.08 & 99.88 $\pm$ 0.03 & 99.91 $\pm$ 0.03 & 99.93 $\pm$ 0.03 \\
Swin & 99.03 $\pm$ 0.06 & 99.91 $\pm$ 0.02 & 99.92 $\pm$ 0.01 & 99.93 $\pm$ 0.00 \\
ViT & 99.27 $\pm$ 0.04 & 99.94 $\pm$ 0.02 & 99.95 $\pm$ 0.01 & 99.96 $\pm$ 0.00 \\
\bottomrule
\end{tabular}
\end{table}

\begin{table}[htbp]
\caption{Flat classification performance after few-shot fine-tuning.}
\label{tab:supp_flat_with_fewshot}
\centering
\small
\begin{tabular}{lcccc}
\toprule
\multicolumn{5}{c}{Vergilius Turonensis} \\
\midrule
Model & Top-1 & Top-5 & Top-10 & Top-30 \\
\midrule
ResNet18 & 68.73 $\pm$ 1.26 & 82.36 $\pm$ 1.13 & 85.68 $\pm$ 0.91 & 89.14 $\pm$ 0.35 \\
ConvNeXt & 79.14 $\pm$ 0.52 & 86.14 $\pm$ 0.32 & 87.32 $\pm$ 0.49 & 89.14 $\pm$ 0.43 \\
Swin & 82.09 $\pm$ 0.69 & 87.59 $\pm$ 0.08 & 89.36 $\pm$ 0.42 & 90.59 $\pm$ 0.41 \\
ViT & 81.77 $\pm$ 0.39 & 88.09 $\pm$ 0.46 & 88.96 $\pm$ 0.20 & 90.55 $\pm$ 0.34 \\
\midrule
\multicolumn{5}{c}{Handwritten} \\
\midrule
Model & Top-1 & Top-5 & Top-10 & Top-30 \\
\midrule
ResNet18 & 97.98 $\pm$ 0.11 & 99.84 $\pm$ 0.01 & 99.91 $\pm$ 0.02 & 99.96 $\pm$ 0.01 \\
ConvNeXt & 99.34 $\pm$ 0.08 & 99.89 $\pm$ 0.02 & 99.92 $\pm$ 0.01 & 99.96 $\pm$ 0.01 \\
Swin & 99.03 $\pm$ 0.07 & 99.92 $\pm$ 0.01 & 99.95 $\pm$ 0.02 & 99.96 $\pm$ 0.00 \\
ViT & 99.32 $\pm$ 0.07 & 99.94 $\pm$ 0.00 & 99.97 $\pm$ 0.00 & 99.97 $\pm$ 0.00 \\
\bottomrule
\end{tabular}
\end{table}

\begin{table}[htbp]
\caption{Hierarchy-aware routing performance on Vergilius Turonensis using the shared tree generated from the trained ResNet18 flat model, without few-shot fine-tuning.}
\label{tab:supp_hierarchy_without_fewshot}
\centering
\scriptsize
\resizebox{\textwidth}{!}{%
\begin{tabular}{llcccc}
\toprule
Model & Level & Top-1 & Top-5 & Top-10 & Top-30 \\
\midrule
ResNet18 & Leaf & 37.59 $\pm$ 0.56 & 46.74 $\pm$ 0.40 & 51.05 $\pm$ 0.94 & 56.45 $\pm$ 0.77 \\
 & Cluster group & 44.76 $\pm$ 0.86 & 54.34 $\pm$ 0.40 & 58.39 $\pm$ 0.98 & 65.41 $\pm$ 0.25 \\
 & SNT group & 44.14 $\pm$ 0.32 & 60.96 $\pm$ 0.99 & 67.68 $\pm$ 0.67 & 77.56 $\pm$ 0.92 \\
\addlinespace
ConvNeXt & Leaf & 36.99 $\pm$ 0.32 & 47.23 $\pm$ 0.45 & 50.00 $\pm$ 0.44 & 54.34 $\pm$ 0.57 \\
 & Cluster group & 40.59 $\pm$ 0.29 & 50.48 $\pm$ 0.39 & 52.88 $\pm$ 0.36 & 57.91 $\pm$ 0.43 \\
 & SNT group & 46.89 $\pm$ 0.43 & 60.81 $\pm$ 0.79 & 65.92 $\pm$ 0.84 & 72.74 $\pm$ 0.38 \\
\addlinespace
Swin & Leaf & 43.78 $\pm$ 0.64 & 54.34 $\pm$ 0.48 & 57.46 $\pm$ 0.92 & 62.07 $\pm$ 0.61 \\
 & Cluster group & 47.79 $\pm$ 0.81 & 57.50 $\pm$ 0.55 & 60.42 $\pm$ 0.83 & 65.29 $\pm$ 0.40 \\
 & SNT group & 54.73 $\pm$ 0.80 & 68.51 $\pm$ 0.71 & 72.56 $\pm$ 0.36 & 79.92 $\pm$ 0.36 \\
\addlinespace
ViT & Leaf & 37.93 $\pm$ 0.49 & 49.81 $\pm$ 1.18 & 53.71 $\pm$ 0.89 & 57.98 $\pm$ 0.69 \\
 & Cluster group & 47.60 $\pm$ 1.21 & 54.76 $\pm$ 0.77 & 57.57 $\pm$ 0.73 & 61.84 $\pm$ 0.62 \\
 & SNT group & 45.80 $\pm$ 0.38 & 61.34 $\pm$ 1.29 & 65.62 $\pm$ 0.67 & 72.93 $\pm$ 0.83 \\
\bottomrule
\end{tabular}%
}
\end{table}

\begin{table}[htbp]
\caption{Hierarchy-aware routing performance on Vergilius Turonensis using the shared tree generated from the trained ResNet18 flat model, after few-shot fine-tuning.}
\label{tab:supp_hierarchy_with_fewshot}
\centering
\scriptsize
\resizebox{\textwidth}{!}{%
\begin{tabular}{llcccc}
\toprule
Model & Level & Top-1 & Top-5 & Top-10 & Top-30 \\
\midrule
ResNet18 & Leaf & 48.27 $\pm$ 1.37 & 60.50 $\pm$ 1.08 & 64.50 $\pm$ 0.98 & 70.82 $\pm$ 0.30 \\
 & Cluster group & 55.68 $\pm$ 1.17 & 66.59 $\pm$ 0.52 & 70.68 $\pm$ 0.33 & 76.27 $\pm$ 0.71 \\
 & SNT group & 53.55 $\pm$ 1.06 & 68.62 $\pm$ 0.41 & 73.91 $\pm$ 1.40 & 82.65 $\pm$ 0.50 \\
\addlinespace
ConvNeXt & Leaf & 71.82 $\pm$ 0.59 & 79.68 $\pm$ 0.15 & 81.73 $\pm$ 0.42 & 85.05 $\pm$ 0.43 \\
 & Cluster group & 74.18 $\pm$ 0.47 & 81.09 $\pm$ 0.36 & 83.00 $\pm$ 0.65 & 85.95 $\pm$ 0.51 \\
 & SNT group & 75.91 $\pm$ 0.52 & 83.83 $\pm$ 0.20 & 85.93 $\pm$ 0.41 & 89.07 $\pm$ 0.34 \\
\addlinespace
Swin & Leaf & 69.91 $\pm$ 0.42 & 78.09 $\pm$ 0.64 & 80.23 $\pm$ 0.35 & 83.04 $\pm$ 0.15 \\
 & Cluster group & 71.00 $\pm$ 0.65 & 79.54 $\pm$ 0.49 & 81.64 $\pm$ 0.13 & 84.45 $\pm$ 0.30 \\
 & SNT group & 74.41 $\pm$ 0.58 & 82.74 $\pm$ 0.27 & 85.39 $\pm$ 0.15 & 89.25 $\pm$ 0.41 \\
\addlinespace
ViT & Leaf & 67.68 $\pm$ 0.30 & 74.50 $\pm$ 0.87 & 76.86 $\pm$ 0.97 & 80.09 $\pm$ 0.45 \\
 & Cluster group & 69.00 $\pm$ 0.20 & 75.46 $\pm$ 0.84 & 78.09 $\pm$ 0.82 & 81.82 $\pm$ 0.29 \\
 & SNT group & 71.54 $\pm$ 0.71 & 78.51 $\pm$ 1.12 & 80.88 $\pm$ 0.83 & 86.52 $\pm$ 0.72 \\
\bottomrule
\end{tabular}%
}
\end{table}

\begin{table}[htbp]
\caption{HD-CNN coarse-to-fine classification performance on Vergilius Turonensis. }
\label{tab:supp_hdcnn}
\centering
\small
\resizebox{\textwidth}{!}{%
\begin{tabular}{lcccccccc}
\toprule
& \multicolumn{4}{c}{Without few-shot fine-tuning} & \multicolumn{4}{c}{With few-shot fine-tuning} \\
\midrule
Model & Top-1 & Top-5 & Top-10 & Top-30 & Top-1 & Top-5 & Top-10 & Top-30 \\
\midrule
ResNet18 & 37.68 $\pm$ 0.39 & 46.78 $\pm$ 0.53 & 50.13 $\pm$ 0.99 & 57.52 $\pm$ 0.57 & 67.21 $\pm$ 0.48 & 76.91 $\pm$ 0.51 & 80.91 $\pm$ 0.53 & 86.43 $\pm$ 0.43 \\
ConvNeXt & 36.53 $\pm$ 0.25 & 45.53 $\pm$ 0.51 & 47.88 $\pm$ 0.31 & 52.27 $\pm$ 0.67 & 66.43 $\pm$ 0.45 & 78.55 $\pm$ 0.65 & 81.45 $\pm$ 0.39 & 86.12 $\pm$ 0.82 \\
Swin & 45.43 $\pm$ 1.17 & 56.82 $\pm$ 0.21 & 60.47 $\pm$ 0.19 & 67.37 $\pm$ 0.31 & 75.76 $\pm$ 0.70 & 82.24 $\pm$ 0.45 & 85.15 $\pm$ 0.73 & 87.94 $\pm$ 0.17 \\
ViT & 43.68 $\pm$ 0.92 & 56.27 $\pm$ 0.70 & 59.62 $\pm$ 0.39 & 63.57 $\pm$ 0.21 & 74.24 $\pm$ 0.48 & 80.49 $\pm$ 0.60 & 82.49 $\pm$ 0.34 & 86.43 $\pm$ 0.56 \\
\bottomrule
\end{tabular}%
}
\end{table}
\FloatBarrier
\subsection{Balanced Accuracy}
\begin{table}[htbp]
\caption{Flat classification balanced accuracy.}
\label{tab:supp_balanced_flat}
\centering
\small
\begin{tabular}{lcccc}
\toprule
& \multicolumn{2}{c}{Without few-shot fine-tuning} & \multicolumn{2}{c}{With few-shot fine-tuning} \\
\midrule
Model & Vergilius & Handwritten & Vergilius & Handwritten \\
\midrule
ResNet18 & 18.85 $\pm$ 1.49 & 98.98 $\pm$ 0.11 & 35.15 $\pm$ 1.75 & 97.99 $\pm$ 0.09 \\
ConvNeXt & 22.83 $\pm$ 1.01 & 99.36 $\pm$ 0.07 & 45.56 $\pm$ 1.54 & 99.35 $\pm$ 0.07 \\
Swin & 29.37 $\pm$ 1.20 & 99.02 $\pm$ 0.06 & 51.13 $\pm$ 0.98 & 99.03 $\pm$ 0.08 \\
ViT & 25.57 $\pm$ 0.80 & 99.29 $\pm$ 0.04 & 50.27 $\pm$ 1.23 & 99.34 $\pm$ 0.07 \\
\bottomrule
\end{tabular}
\end{table}

\begin{table}[htbp]
\caption{Hierarchy-aware routing balanced accuracy with self-generated visual trees.}
\label{tab:supp_balanced_hierarchy_self}
\centering
\small
\begin{tabular}{lcccc}
\toprule
& \multicolumn{2}{c}{Without few-shot fine-tuning} & \multicolumn{2}{c}{With few-shot fine-tuning} \\
\midrule
Model & Vergilius & Handwritten & Vergilius & Handwritten \\
\midrule
ResNet18 & 18.71 $\pm$ 0.22 & 98.98 $\pm$ 0.04 & 22.48 $\pm$ 0.90 & 98.37 $\pm$ 0.05 \\
ConvNeXt & 21.13 $\pm$ 0.77 & 99.12 $\pm$ 0.03 & 33.68 $\pm$ 0.95 & 98.57 $\pm$ 0.07 \\
Swin & 25.62 $\pm$ 0.57 & 98.07 $\pm$ 0.06 & 31.04 $\pm$ 0.97 & 97.16 $\pm$ 0.02 \\
ViT & 19.87 $\pm$ 0.37 & 98.07 $\pm$ 0.06 & 31.67 $\pm$ 0.56 & 97.16 $\pm$ 0.02 \\
\bottomrule
\end{tabular}
\end{table}

\begin{table}[htbp]
\caption{Hierarchy-aware routing balanced accuracy with the shared tree from the trained ResNet18 flat model.}
\label{tab:supp_balanced_shared_baseline_tree}
\centering
\small
\begin{tabular}{lcccc}
\toprule
& \multicolumn{2}{c}{Without few-shot fine-tuning} & \multicolumn{2}{c}{With few-shot fine-tuning} \\
\midrule
Model & Vergilius & Handwritten & Vergilius & Handwritten \\
\midrule
ResNet18 & 18.71 $\pm$ 0.22 & 98.98 $\pm$ 0.04 & 22.48 $\pm$ 0.90 & 98.37 $\pm$ 0.05 \\
ConvNeXt & 18.59 $\pm$ 0.73 & 99.26 $\pm$ 0.05 & 36.85 $\pm$ 0.72 & 99.17 $\pm$ 0.04 \\
Swin & 26.34 $\pm$ 0.35 & 98.81 $\pm$ 0.07 & 38.95 $\pm$ 0.96 & 98.73 $\pm$ 0.04 \\
ViT & 22.13 $\pm$ 1.14 & 99.23 $\pm$ 0.06 & 37.42 $\pm$ 0.82 & 99.10 $\pm$ 0.04 \\
\bottomrule
\end{tabular}
\end{table}

\begin{table}[htbp]
\caption{Hierarchy-aware routing balanced accuracy with the shared tree from ImageNet-pretrained ResNet18.}
\label{tab:supp_balanced_shared_pretrained_tree}
\centering
\small
\begin{tabular}{lcccc}
\toprule
& \multicolumn{2}{c}{Without few-shot fine-tuning} & \multicolumn{2}{c}{With few-shot fine-tuning} \\
\midrule
Model & Vergilius & Handwritten & Vergilius & Handwritten \\
\midrule
ResNet18 & 17.14 $\pm$ 0.71 & 98.75 $\pm$ 0.08 & 21.91 $\pm$ 0.74 & 98.29 $\pm$ 0.09 \\
ConvNeXt & 22.82 $\pm$ 1.25 & 99.17 $\pm$ 0.02 & 37.43 $\pm$ 0.44 & 98.97 $\pm$ 0.02 \\
Swin & 25.17 $\pm$ 1.37 & 98.69 $\pm$ 0.03 & 33.82 $\pm$ 0.23 & 97.99 $\pm$ 0.08 \\
ViT & 20.01 $\pm$ 1.21 & 99.03 $\pm$ 0.03 & 32.62 $\pm$ 0.59 & 98.79 $\pm$ 0.06 \\
\bottomrule
\end{tabular}
\end{table}

\begin{table}[htbp]
\caption{HD-CNN balanced accuracy.}
\label{tab:supp_balanced_hdcnn}
\centering
\small
\begin{tabular}{lcccc}
\toprule
& \multicolumn{2}{c}{Without few-shot fine-tuning} & \multicolumn{2}{c}{With few-shot fine-tuning} \\
\midrule
Model & Vergilius & Handwritten & Vergilius & Handwritten \\
\midrule
ResNet18 & 21.08 $\pm$ 0.83 & 98.94 $\pm$ 0.03 & 36.19 $\pm$ 0.19 & 98.11 $\pm$ 0.09 \\
ConvNeXt & 20.85 $\pm$ 1.14 & 99.19 $\pm$ 0.02 & 34.60 $\pm$ 0.86 & 99.24 $\pm$ 0.05 \\
Swin & 27.30 $\pm$ 0.57 & 98.76 $\pm$ 0.04 & 43.48 $\pm$ 0.46 & 98.89 $\pm$ 0.03 \\
ViT & 23.17 $\pm$ 0.17 & 99.16 $\pm$ 0.04 & 42.76 $\pm$ 0.87 & 99.19 $\pm$ 0.06 \\
\bottomrule
\end{tabular}
\end{table}

\FloatBarrier
\subsection{Additional KNN Retrieval Baselines}
Additional K-nearest-neighbor retrieval baselines are reported to complement the trainable neural baselines. These baselines include five variants. The first variant is a raw pixel-level KNN baseline. Each image is converted to a binarized $64 \times 64$ representation and flattened into a 4096-dimensional vector. The other four variants are feature-based KNN baselines, where feature representations are extracted from the trained ResNet18, ConvNeXt-Tiny, Swin-Tiny, and ViT-Small flat backbones. These variants test whether the learned visual feature spaces are sufficient for nearest-neighbor retrieval without using the trained classifier heads. For all KNN variants, the reference gallery consists of the cleaned SNT standard symbols together with the handwritten training and validation samples. All reference vectors and test vectors are L2-normalized, and cosine similarity is used for nearest-neighbor ranking. For Top-$k$ evaluation, the ranked neighbors are scanned in descending similarity order, and the first $k$ unique class labels are used as candidate predictions. Since KNN is a deterministic retrieval method without trainable classifier parameters, it is not included in the few-shot fine-tuning protocol.

\begin{table}[htbp]
\caption{Additional KNN retrieval baseline performance}
\label{tab:supp_knn}
\centering
\small
\resizebox{\textwidth}{!}{%
\begin{tabular}{lcccccccc}
\toprule
& \multicolumn{4}{c}{Vergilius Turonensis} & \multicolumn{4}{c}{Handwritten} \\
\midrule
Variant & Top-1 & Top-5 & Top-10 & Top-30 & Top-1 & Top-5 & Top-10 & Top-30 \\
\midrule
Pixel-level KNN & 4.20 & 9.42 & 12.17 & 15.80 & 90.12 & 97.95 & 98.43 & 98.93 \\
ResNet18 KNN & 25.19 & 34.93 & 38.68 & 44.98 & 98.52 & 99.94 & 99.96 & 99.96 \\
ConvNeXt KNN & 35.68 & 45.88 & 49.78 & 54.42 & 99.00 & 99.94 & 99.96 & 99.99 \\
Swin KNN & 38.83 & 52.02 & 56.22 & 62.52 & 98.71 & 99.93 & 99.94 & 99.94 \\
ViT KNN & 36.28 & 48.58 & 50.67 & 55.47 & 98.95 & 99.94 & 99.96 & 99.97 \\
\bottomrule
\end{tabular}%
}
\end{table}